\documentclass{IOS-Book-Article}

\usepackage{mathptmx}
\usepackage{soul}\setuldepth{article}
\usepackage{graphicx}
\usepackage{caption}
\usepackage{amsfonts}
\usepackage{wrapfig}
\usepackage{cleveref}
\Crefname{equation}{Eq.}{Eqs.}
\Crefname{figure}{Fig.}{Figs.}
\Crefname{tabular}{Tab.}{Tabs.}
\Crefname{section}{Sec.}{Secs.}

\def\hb{\hbox to 11.5 cm{}}

\begin{document}

\pagestyle{headings}
\def\thepage{}

\begin{frontmatter}              

\title{Less is More: A Call to Focus on Simpler Models in Genetic Programming for Interpretable Machine Learning}

\markboth{}{}

\author[A]{\fnms{Marco} \snm{Virgolin}%
\thanks{Corresponding Author: Marco Virgolin, Centrum Wiskunde \& Informatica, Science Park 123,
1098 XG Amsterdam, The Netherlands; E-mail:
marco.virgolin@cwi.nl.}},
\author[B]{\fnms{Eric} \snm{Medvet}}, 
\author[C]{\fnms{Tanja} \snm{Alderliesten}}
and
\author[A]{\fnms{Peter A.N.} \snm{Bosman}}


\address[A]{Centrum Wiskunde \& Informatica, Amsterdam, The Netherlands}
\address[B]{University of Trieste, Trieste, Italy}
\address[C]{Leiden University Medical Center, Leiden, The Netherlands}

\begin{abstract}
Interpretability can be critical for the safe and responsible use of machine learning models in high-stakes applications.
So far, evolutionary computation (EC), in particular in the form of genetic programming (GP), represents a key enabler for the discovery of interpretable machine learning (IML) models.
In this short paper, we argue that research in GP for IML needs to focus on searching in the space of low-complexity models, by investigating new kinds of  search strategies and recombination methods.
Moreover, based on our experience of bringing research into clinical practice, we believe that research should strive to design better ways of modelling and pursuing interpretability, for the obtained solutions to ultimately be most useful.
\end{abstract}

\end{frontmatter}

\section{Background}
The field of explainable artificial intelligence (XAI) makes important contributions toward meeting the challenge of fairness and accountability in AI.
The goal of XAI is to design algorithms that can help explain the decisions taken by unintelligible models (e.g., deep neural networks) or that can generate inherently interpretable machine learning (IML) models~\cite{adadi2018peeking,arrieta2020explainable}.
Evolutionary computation (EC) is routinely used as a powerful engine for both cases, see, e.g.,~\cite{johnson2000explanatory,cano2013interpretable,virgolin2020explaining,lensen2020genetic,dandl2020multi,virgolin2022robustness}.
Here, we focus on how EC is often used to generate IML models.
A prime example of this is embodied by genetic programming (GP)~\cite{koza1994genetic} (or similar algorithms, such as grammatical evolution~\cite{o2001grammatical}) for symbolic regression (SR), i.e., techniques inspired by natural evolution to discover a governing equation from data through such means as selection and recombination~\cite{schmidt2009distilling}.
GP and similar techniques have also been applied to other types of problems than SR (see, e.g., \cite{virgolin2020explaining,lensen2020genetic,espejo2009survey,hein2018interpretable,de2020genetic}); using other types of \emph{symbolic models}, i.e., models composed of high-level operations such as arithmetic operations, Boolean conditions, software instructions (see, e.g.,~\cite{cano2013interpretable,mansoori2008sgerd,aitkenhead2008co,zhang2022pstree}).

\section{Open problems and possible directions in GP for IML}
\begin{figure}
    \centering
    \begin{tabular}{cc}
        \includegraphics[width=0.44\linewidth]{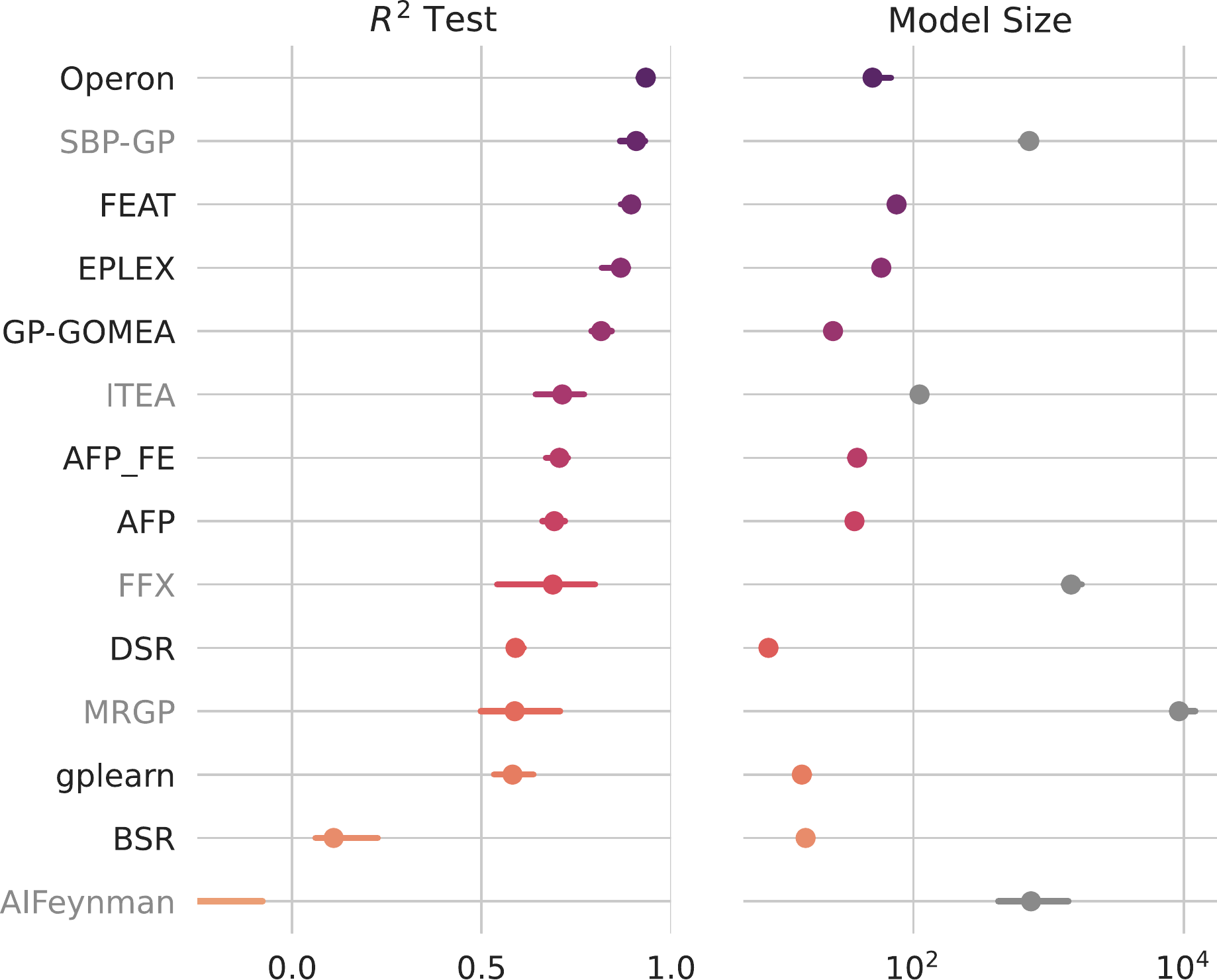} & \includegraphics[width=0.36\linewidth]{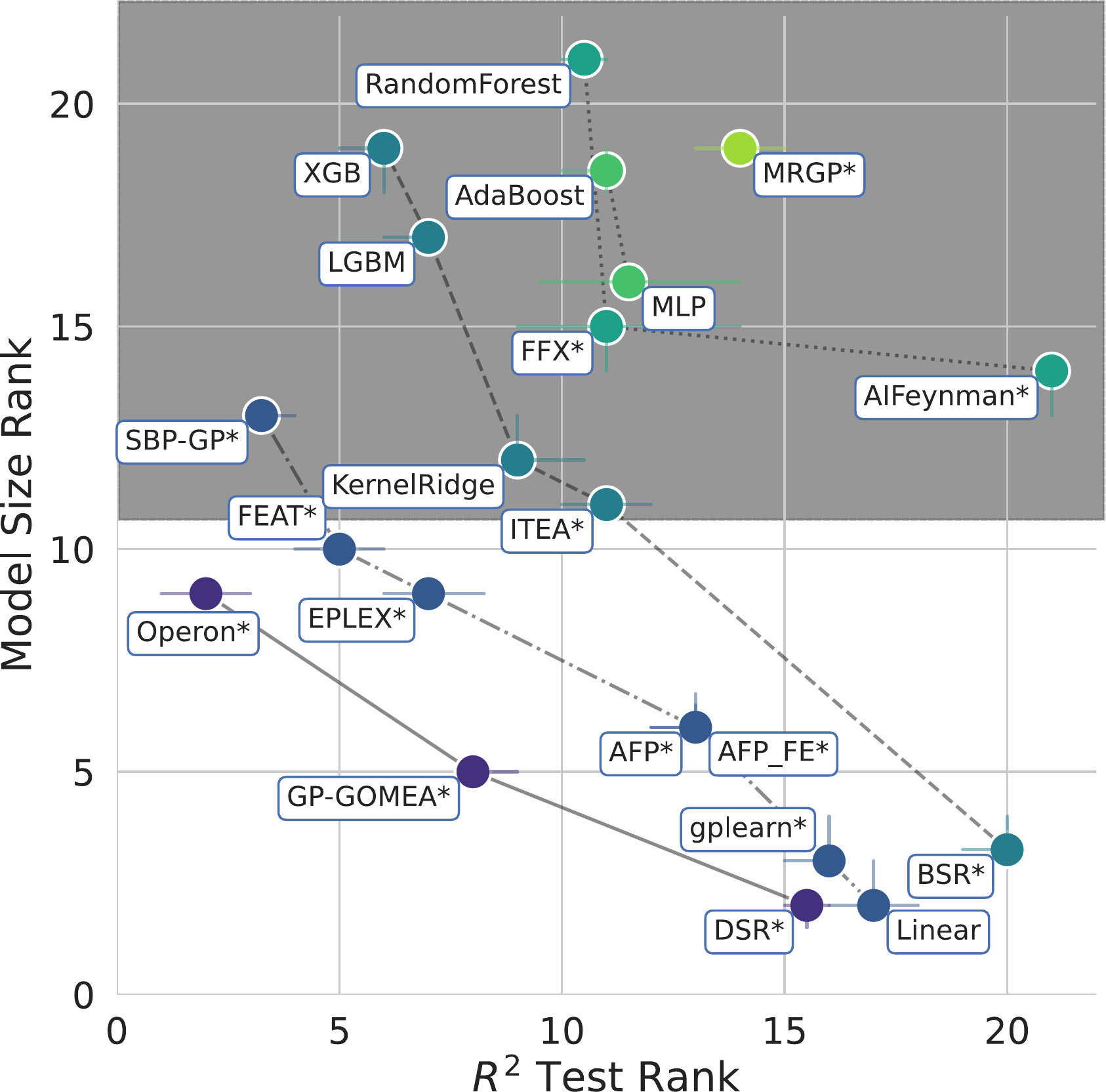} 
    \end{tabular}
    \caption{
    Algorithms in gray (left) or in the gray area (right) produce models with $>100$ components on average on SRBench v.2.0.
    \textbf{Left:} Pairgrid of model accuracy (measured in terms of $R^2$ on the test set) and model size (measured in terms of total number of components) for the symbolic algorithms of SRBench. 
    \textbf{Right:} Trade-off fronts in terms of model accuracy and size ranks for all the algorithms of SRBench (here, symbolic algorithms are starred).
    Plots made with~\cite{srbench}.
    }
    \label{fig:srbench}
\end{figure}

A large drive in GP is to improve model accuracy much akin to a key performance indicator in ML.
Unfortunately, sometimes this is done at the cost of ignoring the applicability of the approach.
In fact, this has pushed research towards designing GP algorithms that, striving for maximizing accuracy, end up delivering models with hundreds, thousands, or even more components~\cite{martins2018solving}.
If we consider the current results reported in SRBench (v.2.0), a large benchmark on SR~\cite{la2021contemporary}, a number of the competing algorithms produce models with more than $100$ components; see \Cref{fig:srbench}.
Based on our experience, even if composed of high-level operations, such models are already too large to be interpretable~\cite{virgolin2021improving}.
In our view, GP research aimed at IML should strive to improve the discovery of accurate models, subject to these models having limited complexity.
Even so, the use of GP is likely to remain justified, as the number and type of operations that can be considered still leads to large search spaces that lack linearity or convexity. 
Searching for low-complexity models therefore calls for better search strategies. In fact, typical recombination methods (e.g., subtree crossover and mutation in GP~\cite{koza1994genetic}) but also new ones (e.g., semantic recombination~\cite{moraglio2012geometric,pawlak2014semantic}) replace or stack entire (and often large) blocks of components, leading to model growth~\cite{vanneschi2010measuring}, and are ineffective when the models at play need to remain small~\cite{virgolin2021improving,liu2022evolvability}.
We are attempting to answer this need by studying how pattern identification and propagation studied in classic genetic algorithms~\cite{thierens2011optimal} can be incorporated in GP: the result of our efforts so far, GP-GOMEA, achieves some of the best trade-offs between model accuracy and simplicity on SRBench (see \Cref{fig:srbench}).

Besides this, we believe that better objectives that act as proxies for interpretability are needed.
Literature research typically adopts simplistic metrics as proxies for interpretability, such as model size or hand-crafted scores (see, e.g.,~\cite{hein2018interpretable,vladislavleva2008order}).
In our experience in clinical applications~\cite{maree2019evaluation,virgolin2020surrogate,van2021robust}, obtaining a good objective function of what the user truly needs requires several sessions of interaction.
A promising direction here is to use ML itself to learn what specific users find to be more or less interpretable in an automatic fashion~\cite{virgolin2020learning,virgolin2021model}.
Moreover, researchers should strive to include user studies, to assess whether the proposed objective (interpretability, trust, etc.) obtains the desired effect in practice~\cite{virgolin2021model}.

\bibliography{main}
\end{document}